\documentclass[letterpaper, 10 pt, journal]{IEEEtran}

\IEEEoverridecommandlockouts  % This command is only needed if you want to use the \thanks command

\usepackage{fix-cm}
\usepackage{etex}

\usepackage{dblfloatfix}

\usepackage{nag}

\makeatletter
\@ifpackageloaded{xcolor}{}{%
\usepackage[table,x11names,dvipsnames,svgnames]{xcolor}%
}
\makeatother

\usepackage{colortbl}

\usepackage{graphicx}
\usepackage{wrapfig}

\definecolorset{RGB}{lyft}{}{Red,194,39,36;Sunset,202,53,33;Orange,205,68,20;Amber,200,117,42;Yellow,242,169,52;Citron,186,188,44;Lime,112,159,33;Green,56,139,31;Mint,45,118,56;Teal,52,133,135;Cyan,60,132,202;Blue,55,94,248;Indigo,64,13,247;Purple,115,42,248;Pink,176,25,145;Rose,176,32,75}

\definecolorset{HTML}{h}{}{grapefruit,ED5565;bittersweet,FC6E51;sunflower,FFCE54;grass,A0D468;mint,48CFAD;aqua,4FC1E9;bluejeans,5D9CEC;lavender,AC92EC;pinkrose,EC87C0;lightgray,F5F7FA;mediumgray,CCD1D9;darkgray,656D78}

\usepackage{cite}

\usepackage{microtype}

\usepackage[american]{babel}

\usepackage{array}
\usepackage{multirow}
\usepackage{booktabs}
\usepackage{makecell} %introduces \thead and \makecell

\ifcsname labelindent\endcsname

\fi
\usepackage[inline]{enumitem}

\usepackage{subfig}

\newenvironment{lenumerate}[2][]
{\begin{enumerate}[label=(#2\arabic*),leftmargin=0.2in,itemindent=0.15in,#1]}
{\end{enumerate}}

\setlist*[enumerate,1]{label={\itshape\arabic*)}}

\makeatletter
\newcommand{\paragraphswithstop}{%
\let\copyparagraph\paragraph%
\renewcommand\paragraph[1]{\copyparagraph{##1.}}%
}
\makeatother

\usepackage[framemethod=tikz]{mdframed}

\makeatletter
\def\namedlabel#1#2{\begingroup
  #2%
  \def\@currentlabel{#2}%
  \phantomsection\label{#1}\endgroup
}
\makeatother

\makeatletter
\def\namedlabelphantom#1#2{\begingroup
  \def\@currentlabel{#2}%
  \phantomsection\label{#1}\endgroup
}
\makeatother

\newcommand{\parunskip}{\bgroup\unskip\parfillskip=0pt \par\egroup}

\input{preamble/math}
\usepackage{units}

  \newcommand{\newcolorlabel}[2]{%
  \expandafter\newcommand\csname #1\endcsname[1]{%
    \tikz[baseline]{\node[text=white,fill=#2,anchor=base,text height=1.3ex,text depth=0.1ex,font=\sffamily\bfseries]{##1}}}%
}

\newcommand{\newcommenter}[2]{%
  \expandafter\newcommand\csname #1\endcsname[1]{%
    \fcolorbox{#2}{#2}{\color{white}\textsf{\textbf{#1}}}
    {\color{#2}##1}}%
  \expandafter\newcommand\csname at#1\endcsname{%
    \fcolorbox{#2}{#2}{\color{white}\textsf{\textbf{@#1}}}
    {\color{#2}}}%
  \expandafter\newcommand\csname #1cite\endcsname[1]{%
    \csname #1\endcsname {[##1]}
  }%
  \expandafter\newcommand\csname #1ref\endcsname[1]{%
    \csname #1\endcsname {$\blacktriangleright$##1}
  }%
  \expandafter\newcommand\csname #1hl\endcsname[2]{%
    \colorbox{#2}{\color{white}\textsf{\textbf{#1}}}\sethlcolor{Azure2}\hl{##2}~%
    \expandafter\ifx\csname commentarrow\endcsname\relax$\leftarrow$\else \commentarrow[#2]\fi~%
    {\color{#2}##1}}%
  \expandafter\newcommand\csname #1st\endcsname[2]{%
    \colorbox{#2}{\color{white}\textsf{\textbf{#1}}}\sout{##2}~%
    \expandafter\ifx\csname commentarrow\endcsname\relax$\leftarrow$\else \commentarrow[#2]\fi~%
    {\color{#2}##1}}%
}
\newcommenter{TODO}{DodgerBlue1}
\newcommenter{rtron}{Green3}

\usepackage{comment}

\usepackage{pdfcomment}

\usepackage{soul}

\usepackage[normalem]{ulem}

\usepackage{csquotes}

\usepackage{suffix}

\usepackage{environ}

\makeatletter
\newsavebox{\boxifnotempty}
\newcommand{\displayifnotempty}[3]{\sbox\boxifnotempty{#2}\setbox0=\hbox{\usebox{\boxifnotempty}\unskip}%
  \ifdim\wd0=0pt
  \else
  #1\usebox{\boxifnotempty}#3%
  \fi%
}

\newcommand{\ifempty}[2]{\setbox0=\hbox{#1\unskip}%
  \ifdim\wd0=0pt%
  #2%
  \fi%
}

\newcommand{\ifnotempty}[2]{\setbox0=\hbox{#1\unskip}%
  \ifdim\wd0>0pt%
  #2%
  \fi%
}
\makeatother

\newcommand{\switchifempty}[3]{\sbox\boxifnotempty{#1}\setbox0=\hbox{\usebox{\boxifnotempty}\unskip}%
  \ifdim\wd0=0pt{}%
  #2%
  \else{}%
  #3%
  \usebox{\boxifnotempty}%
  \fi%
}

\makeatletter
\@ifundefined{chapter}{\usepackage{algorithm}}{\usepackage[chapter]{algorithm}}
\makeatother
\usepackage{algorithmicx}
\usepackage{algpseudocode}
\makeatother%

\usepackage{scrlfile}

\makeatletter
\newcommand*\newstoreddef[1]{
  \BeforeClosingMainAux{%
    \immediate\write\@auxout{%
      \string\restoredef{#1}{\csname #1\endcsname}%
    }%
  }%
}
\newcommand*{\restoredef}[2]{% used at the aux file
  \expandafter\gdef\csname stored@#1\endcsname{#2}%
}
\newcommand*{\storeddef}[1]{
  \@ifundefined{stored@#1}{0}{\csname stored@#1\endcsname}%
}
\makeatother

\usepackage{pageslts}
\NewEnviron{tee}{\BODY\typeout{Marker Tee [start] ^^J \BODY ^^JMaker Tee [end]}}

\usepackage{cleveref}
\usepackage{nameref}

\usepackage{tikz}
\usetikzlibrary{calc}
\usetikzlibrary{matrix}
\usetikzlibrary{chains,scopes}
\usetikzlibrary{shapes.geometric}
\usetikzlibrary{arrows.meta}
\usetikzlibrary{decorations.markings}
\usetikzlibrary{decorations.pathreplacing}
\usetikzlibrary{backgrounds}

\tikzset{
  dim above/.style={to path={\pgfextra{
        \pgfinterruptpath
        \draw[>=latex,|->|] let
        \p1=($(\tikztostart)!1.5em!90:(\tikztotarget)$),
        \p2=($(\tikztotarget)!1.5em!-90:(\tikztostart)$)
        in(\p1) -- (\p2) node[pos=.5,sloped,above]{#1};
        \endpgfinterruptpath
      }
    }
  },
  dim double above/.style={to path={\pgfextra{
        \pgfinterruptpath
        \draw[>=latex,|->|] let
        \p1=($(\tikztostart)!3em!90:(\tikztotarget)$),
        \p2=($(\tikztotarget)!3em!-90:(\tikztostart)$)
        in(\p1) -- (\p2) node[pos=.5,sloped,above]{#1};
        \endpgfinterruptpath
      }
    }
  },
  dim below/.style={to path={\pgfextra{
        \pgfinterruptpath
        \draw[>=latex,|->|] let
        \p1=($(\tikztostart)!-1em!-90:(\tikztotarget)$),
        \p2=($(\tikztotarget)!-1em!90:(\tikztostart)$)
        in (\p1) -- (\p2) node[pos=.5,sloped,below]{#1};
        \endpgfinterruptpath
      }
    }
  },
}

\tikzset{
    right angle quadrant/.code={
        \pgfmathsetmacro\quadranta{{1,1,-1,-1}[#1-1]}     % Arrays for selecting quadrant
        \pgfmathsetmacro\quadrantb{{1,-1,-1,1}[#1-1]}},
    right angle quadrant=1, % Make sure it is set, even if not called explicitly
    right angle length/.code={\def\rightanglelength{#1}},   % Length of symbol
    right angle length=2ex, % Make sure it is set...
    right angle symbol/.style n args={3}{
        insert path={
            let \p0 = ($(#1)!(#3)!(#2)$) in     % Intersection
                let \p1 = ($(\p0)!\quadranta*\rightanglelength!(#3)$), % Point on base line
                \p2 = ($(\p0)!\quadrantb*\rightanglelength!(#2)$) in % Point on perpendicular line
                let \p3 = ($(\p1)+(\p2)-(\p0)$) in  % Corner point of symbol
            (\p1) -- (\p3) -- (\p2)
        }
    }
}

\newcommand{\pgfextractangle}[3]{%
    \pgfmathanglebetweenpoints{\pgfpointanchor{#2}{center}}
                              {\pgfpointanchor{#3}{center}}
    \global\let#1\pgfmathresult
}

\usetikzlibrary{shapes.arrows}
\newcommand{\commentarrow}[1][Azure4]{\tikz[baseline=-3pt]{\node[shape border uses incircle, fill=#1,rotate=180,single arrow, inner sep=1pt, minimum size=6pt, single arrow head extend=2pt]{};}}

\tikzset{ax/.style={-latex,line width=2pt}}

\tikzset{camera/.style={fill=Sienna1,fill opacity=0.5},%
image plane/.style={draw=RoyalBlue3,line width=2pt}}

\usepackage{subfig}
\usepackage{algorithm}
\usepackage{algpseudocode}
\usepackage{makecell}

\usepackage{xcolor}
\usepackage{pifont}

\newcommand{\cmark}{\textcolor{green}{\ding{51}}}
\newcommand{\xmark}{\textcolor{red}{\ding{55}}}

\makeatletter
\newcommand{\@LN@col}[1]{}
\newcommand{\@LN}[2]{}
\makeatother

\newcommenter{todo}{Firebrick1}
\newcommenter{dawei}{Orange}
\newcommenter{nuo}{blue}
\newcommenter{zhiwen}{green}

\usepackage{array}
\usepackage{graphicx}
\usepackage{hyperref}
\hypersetup{hidelinks}

\title{\LARGE \bf
Embedding Semantic Risk into Distance Fields and CBFs \\ for Online Monocular Safe Control

}

\author{Dawei Zhang$^{*1}$, Nuo Chen$^{*3}$, Shuo Liu$^{2}$, Roberto Tron$^{1}$ and Zhiwen Fan$^{3}$
\thanks{$^*$Denotes equal contribution.}
\thanks{$^{1}$Dawei Zhang and Roberto Tron are with the Division of Systems Engineering,
        Boston University, United States
        {\tt\small (dwzhang, tron)@bu.edu}}%
\thanks{$^{2}$ Shuo Liu is with the  Department of Mechanical Engineering,
        Boston University, United States
        {\tt\small (liushuo)@bu.edu}}%      
\thanks{$^{3}$ Nuo Chen and Zhiwen Fan are with the Department of Electrical and Computer Engineering at Texas A\&M University, United States
        {\tt\small (nuochen, zhiwenfan)@tamu.edu}}%
\thanks{This work has been submitted to the IEEE for possible publication. Copyright may be transferred without notice, after which this version may no longer be accessible.}
}

\IEEEaftertitletext{%
\vspace{-0.45em}
\centerline{\textbf{Project Website:}\quad\url{https://safedf.github.io/}}
\vspace{0.45em}
}

\begin{document}

\maketitle

\thispagestyle{empty}
\pagestyle{empty}

%%%%%%%%%%%%%%%%%%%%%%%%%%%%%%%%%%%%%%%%%%%%%%%%%%%%%%%%%%%%%%%%%%%%%%%%%%%%%%%%
\begin{abstract}
We propose an online monocular perception-to-control framework that embeds semantic risk into the distance field used by Control Barrier Function (CBF)-based safe navigation and teleoperation. Many perception-based safety filters assign the same distance-based safety margin to all mapped obstacles or use semantics only as a downstream controller adjustment, rather than encoding semantic risk in the spatial representation. Our framework instead reasons online about obstacle geometry and class-dependent risk by embedding semantic information directly into the Euclidean Signed Distance Field (ESDF). This design encodes semantic risk before control optimization, so high-risk objects exert a larger spatial influence in the safety field while retaining efficient ESDF queries at runtime. Specifically, a foundation-model-based SLAM front end reconstructs dense 3-D geometry from monocular RGB video, while per-frame semantic segmentation provides pixel-level class labels that are fused into the reconstructed geometry. The resulting geometric-semantic representation is then converted into an ESDF, where semantic labels identify safety-relevant regions and impose class-dependent inflation before field computation. The semantic-aware ESDF provides the local distance values and spatial derivatives required by the CBF controller, while class-dependent gains further regulate the controller response. Extensive simulation and hardware experiments demonstrate online operation at 10--20 Hz and semantic-aware safe behavior in both teleoperation and autonomous navigation.

\end{abstract}

%%%%%%%%%%%%%%%%%%%%%%%%%%%%%%%%%%%%%%%%%%%%%%%%%%%%%%%%%%%%%%%%%%%%%%%%%%%%%%%%
\section{INTRODUCTION}

Robotic navigation and teleoperation depend on a tight perception-to-control loop that can operate online in cluttered environments. With carefully designed sensing hardware, known camera models, or depth input, dense geometric maps can be constructed and used for safety filtering. However, for a lightweight monocular setup, building a dense and control-compatible safety representation from RGB video remains difficult. Safe behavior also depends not only on where obstacles are, but on what they are. A person, a pet, and a cardboard box may occupy spaces with similar geometry while calling for different safety behaviors. Control Barrier Functions (CBFs) provide a principled framework for enforcing safety in robotic systems \cite{AmesTAC2017,Ames2019,Xiao2019,liu2023auxiliary}, but their effectiveness depends on how scene observations are converted into a representation that can be queried by the controller.

Early CBF-based safe control methods typically assume that the environment is known or can be represented by simple geometric obstacles, such as circles, spheres, or other analytical sets
\cite{AmesTAC2017,Ames2019,Long2021}. These formulations provide clean safety constraints, but their reliance on manually specified obstacle geometry limits their applicability in online robotic deployment.
% \textcolor{red}{I do not understand, what's the connection between previous sentences with the following sentences, and how to introduce our method given the following sentences (we also estimate geometry via perception, just we are using foundation model, and ESDF)}
More recent work extends CBF-based safe control to environments whose geometry is estimated online from perception. In these methods, obstacle representations derived from point clouds, radiance fields, occupancy maps, Gaussian splatting maps, or other sensor observations are used to construct distance-based CBF constraints \cite{tong2023_nerfs_cbf,sa_2024_pointcloud_cbf,Zhou2024,chen2024safer-splat,Long2021}. These studies show that perception can be coupled with CBF-based safety enforcement, but the resulting constraints remain largely geometry-driven. Once an obstacle is mapped, the controller mainly reacts to distance, so high-risk entities and benign clutter are often treated with the same level of conservativeness.

One way to address this geometry-only limitation is to use semantics when defining the safety filter. Existing semantic CBF methods use labels or scene context to adjust safety margins or controller conservativeness \cite{qian2024semantic,sanyal2025asma,ChenChandra2026}. However, in most cases, semantics are added after the geometric map has already been built, typically through risk scores or controller parameters. The distance field queried by the CBF therefore remains mostly geometry-defined, and obstacle identity does not directly change the spatial support from which safety constraints are computed. Many of these systems also rely on additional depth measurements or dense mapping front ends with stronger sensing assumptions, which makes them less suitable for monocular deployment.

\begin{figure}[t]
    \centering
    \includegraphics[width=\linewidth]{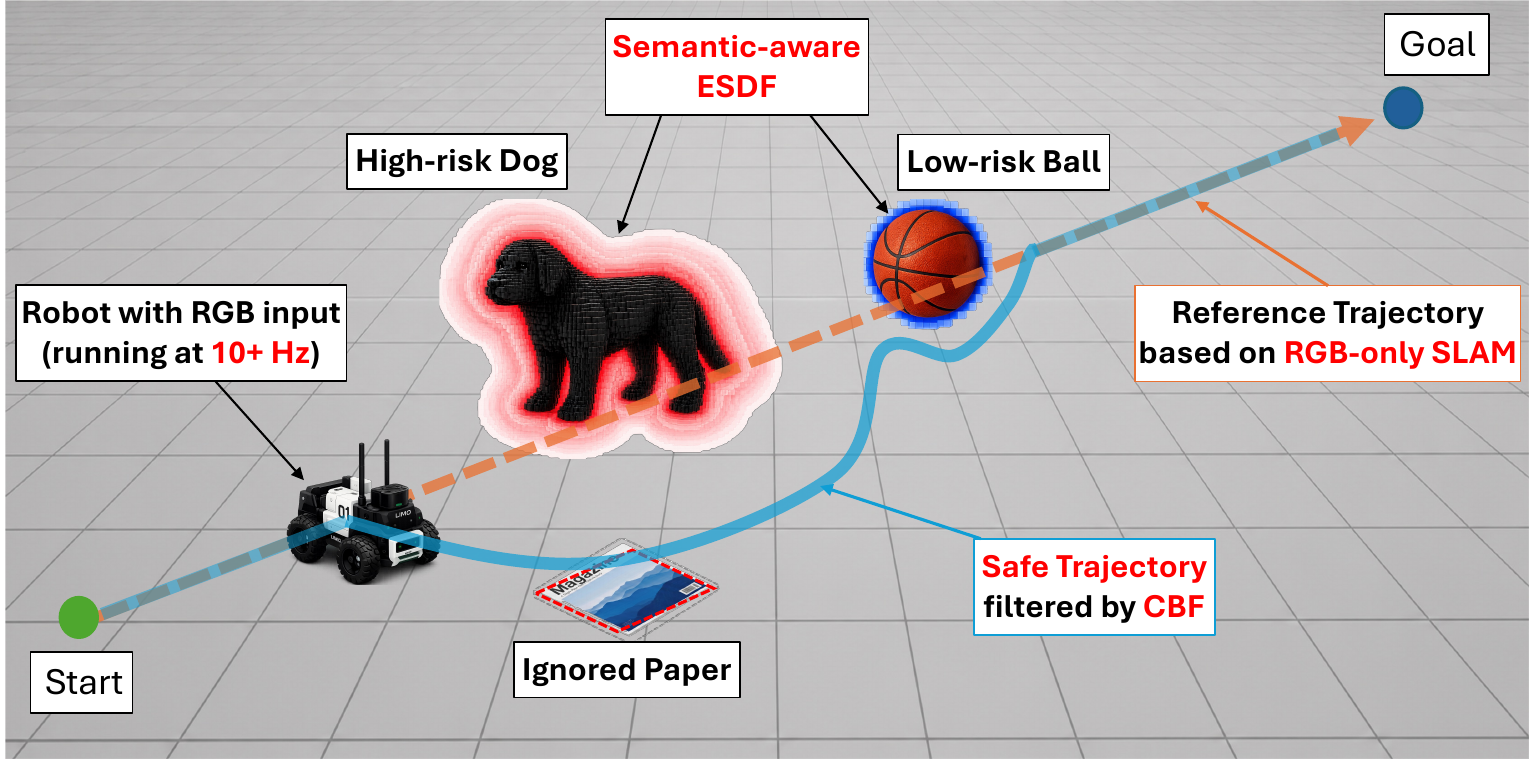}
    \caption{\small Online semantic-aware safe navigation based on monocular dense SLAM, semantic mapping, and CBFs.}
    \label{fig:overview}
\vspace{-5mm}
\end{figure}

Recent advances in foundation model (FM)-based visual reconstruction, such as DUSt3R, MASt3R, and VGGT, provide a new way to recover dense geometry from monocular visual observations \cite{wang2024dust3r,leroy2024grounding,wang2025vggt,murai2025mast3r,maggio2026vggtslam2}. These models use learned 3-D reconstruction priors to recover dense geometry from images, and FM-based SLAM extends this capability to online monocular video, including settings without a fixed or parametric camera model for visual reconstruction.
Table~\ref{tab:capability_comparison} summarizes the trade-offs among
existing SLAM representations. FM-based monocular SLAM offers a suitable
front end for CBF-based control by providing dense geometry at an online
rate, allowing safety constraints to be built from distance values and
spatial derivatives queried at control time.
% As shown in Table~\ref{tab:capability_comparison}, sparse SLAM supports online monocular mapping but provides limited geometry for safety-field construction, while conventional dense SLAM often depends on depth measurements, known camera models, or offline multi-view pipelines.
% FM-based monocular SLAM offers a suitable front end for CBF-based control by providing dense geometry at an online rate, allowing safety constraints to be built from distance values and spatial derivatives queried at control time.

However, this geometry alone does not define semantic safety: the map does not specify which scene regions are safety-relevant or how obstacle identity should shape the field queried by the controller. Motivated by this gap, we propose a semantic-aware safety filtering framework that integrates FM-based monocular dense SLAM with CBFs for online navigation and teleoperation, as illustrated in Fig.~\ref{fig:overview}. The reconstructed geometry is fused with per-frame semantic labels and converted into a semantic-aware ESDF, where labels determine obstacle
selection and class-dependent inflation before field computation. The resulting ESDF is used to construct CBF constraints, while class-dependent CBF gains regulate the dynamic response of the safety filter. This unified pipeline gives higher-risk objects a larger spatial
influence and more conservative CBF response.

The main contributions of this work are summarized as follows:
\begin{itemize}
\item An online monocular perception-to-control framework that turns FM-based dense SLAM from RGB video into a control-ready safety field, enabling CBF-based navigation and teleoperation under a lightweight visual sensing setup.

\item A semantic-aware ESDF formulation that embeds obstacle identity into both field construction and CBF design. Semantic labels select safety-relevant geometry and set class-dependent inflation before distance-field computation, while class-dependent CBF gains adjust the safety filter response at control time.

\item Extensive validation in simulation and hardware experiments, showing that the proposed framework preserves online efficiency, runs at 10--20 Hz, and produces semantic-aware safety behavior in both teleoperation and autonomous navigation.
\end{itemize}

\begin{table}[t]
\centering
\caption{\small Capability comparison of mapping approaches for online CBF-based safety control.}
\label{tab:capability_comparison}
\setlength{\tabcolsep}{3pt}
\renewcommand{\arraystretch}{1.12}
\resizebox{\columnwidth}{!}{
\begin{tabular}{lccccc}
\toprule
Approach 
& \makecell{Online\\Mapping}
& \makecell{Dense\\Geometry}
& \makecell{Monocular\\Input}
& \makecell{Calibration\\Free}
& \makecell{Semantic\\Safety} \\
\midrule
Sparse SLAM 
& \cmark & \xmark & \cmark & \xmark & \xmark \\
Dense SLAM
& \xmark & \cmark & \xmark & \xmark & \xmark \\
FM-based SLAM 
& \cmark & \cmark & \cmark & \cmark & \xmark \\
Ours
& \cmark & \cmark & \cmark & \cmark & \cmark \\
\bottomrule
\end{tabular}
}
\vspace{-3mm}
\end{table}

\subsection{Related Work}
\textbf{Vision-based foundation models.}
Recent progress in vision foundation models has substantially improved RGB-only 3-D perception. 
CroCo learns geometry-aware visual representations through cross-view pretraining, providing strong priors for downstream 3-D understanding tasks~\cite{weinzaepfel2022croco}. 
Building on this direction, DUSt3R reformulates dense 3-D reconstruction as pointmap prediction from image pairs, enabling calibration-free geometric estimation directly from visual observations~\cite{wang2024dust3r}. 
MASt3R further improves geometric matching and reconstruction robustness under challenging viewpoint changes~\cite{leroy2024grounding}. 
More recently, MASt3R-SLAM extends these learned geometric priors to online monocular dense SLAM, enabling joint pose estimation and dense scene reconstruction from RGB input~\cite{murai2025mast3r}. These methods provide the dense RGB-based geometric foundation needed for our setting. Our work builds on this capability by fusing such geometry with semantics and converting it into a control-compatible
safety representation for online CBF-based control.

\textbf{Perception-based CBFs.}
A growing body of work integrates perception-derived geometry with CBFs for safe navigation. Early neural radiance fields methods reconstruct scene geometry offline for CBF constraint construction, but their reliance on photometric optimization makes online map updates difficult~\cite{tong2023_nerfs_cbf}. Point-cloud-based methods construct distance-based CBF constraints directly from local observations, e.g., by fitting surface models to raw point sets to obtain analytical barriers~\cite{sa_2024_pointcloud_cbf}.
They support reactive safety filtering without global mapping, but the
resulting constraints remain local, frame-dependent, and sensitive to
instantaneous depth quality. Gaussian Splatting (GS) can provide an online scene representation from which obstacle approximations are derived for CBF constraints~\cite{chen2024safer-splat}. Semantic cues have also been fused with geometric reconstruction, such as TSDF maps, in MPC-CBF frameworks for navigation in semi-static environments~\cite{qian2024semantic}. In teleoperation, visual-inertial SLAM has been combined with CBF-based safety filtering in complex environments~\cite{Zhou2024}. These works demonstrate the practical value of coupling online perception, mapping, and safety control, but the safety representations remain primarily geometry-driven, with limited integration of semantic information into the geometric field used for CBF evaluation.

Complementary efforts use semantics or high-level context to improve safety-aware behavior. Vision-language models have been used to adjust the conservativeness of CBF-based controllers through risk-aware parameter modulation~\cite{sanyal2025asma,ChenChandra2026}. In contrast, our framework integrates semantics directly into the safety pipeline: semantic labels are used both to select safety-relevant obstacles and to shape the ESDF through class-dependent margins, while the resulting semantic-aware ESDF is used to construct the CBF constraints.

% Motivated by this, we propose a semantic-aware safety filtering framework that integrates FM-based monocular dense SLAM with CBFs for online navigation and teleoperation, as illustrated in Fig.~\ref{fig:overview}. The key idea is to encode semantic risk directly in the online spatial representation used by the controller, rather than introducing semantics as a separate heuristic after mapping.
% Within this pipeline, the ESDF plays a central role. It converts reconstructed geometry into a control-compatible field that provides continuous distance queries and local gradients for optimization-based control. We further shape this field with class-dependent safety margins, so semantic information modifies the safe geometry itself before barrier constraints are formed. CBF constraints are then constructed directly from the semantically shaped ESDF, yielding a unified perception-to-control pipeline in which higher-risk objects trigger earlier and more conservative intervention through the same control mechanism. The resulting perception-to-control pipeline supports efficient online safety filtering and achieves interactive control rates of 10--20 Hz in our implementation.

\section{Preliminaries}
\begin{figure*}[t] 
    \centering
    \includegraphics[width=\linewidth]{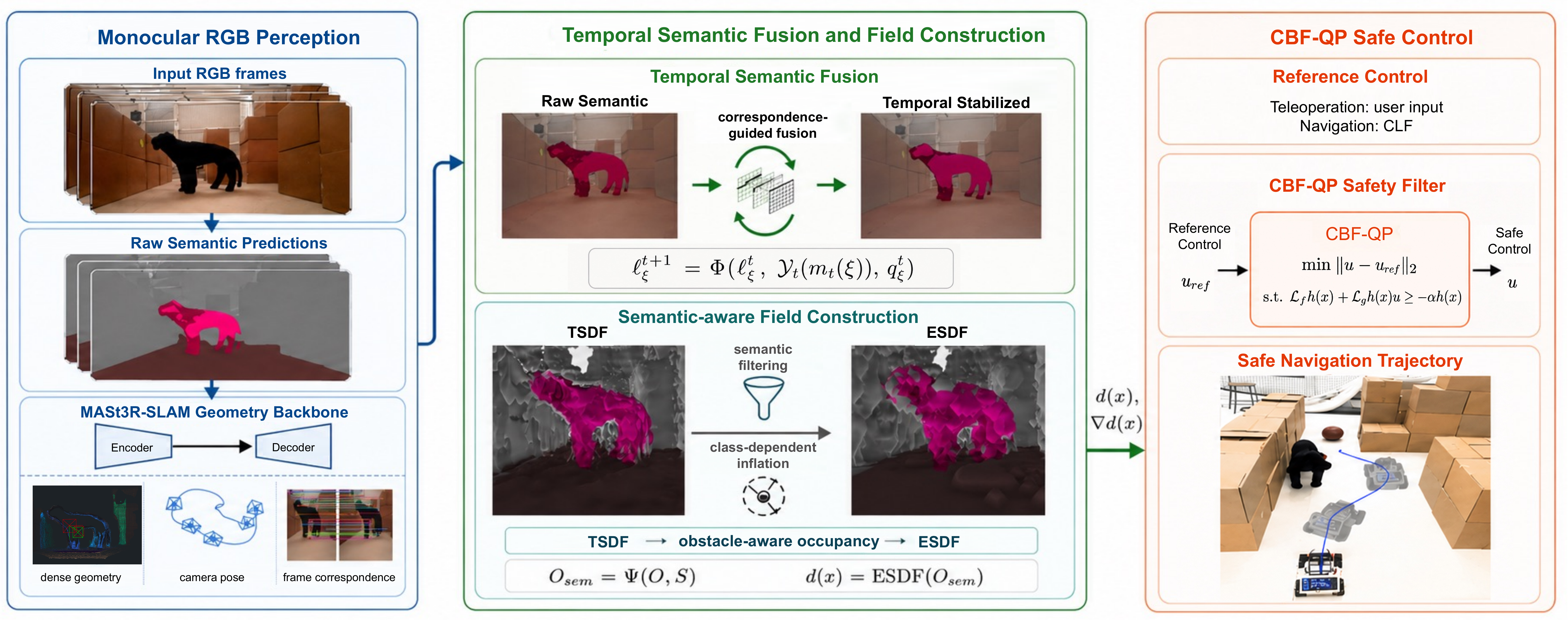}
\caption{\small Overview of the proposed online semantic-aware safe control framework.
Monocular RGB frames are processed by semantic segmentation and MASt3R-SLAM-based dense geometry estimation. Semantic labels are temporally fused with reconstructed 3-D geometry, which is integrated into a local TSDF and converted into obstacle-aware occupancy before ESDF construction. Obstacle filtering and class-dependent inflation encode risk directly into the distance field. The resulting distance and gradient are used by a CBF-QP safety filter to minimally adjust the reference control and generate safe navigation or teleoperation commands.}
    \label{fig:system_overview}
\vspace{-3mm}
\end{figure*}

\subsection{MASt3R-SLAM}
We build our system upon the foundation-model-based MASt3R-SLAM framework \cite{murai2025mast3r}, which performs dense 3-D reconstruction and pose estimation from monocular RGB images by exploiting learned geometric priors. Its capability to generate consistent, metric-scale geometry from vision-only inputs makes it particularly suitable for our setting, as it allows us to construct a local SDF that can be further enriched with semantic information and directly embedded into CBF constraints.

Given an incoming RGB frame $I_t$ and a keyframe $I_k$,
the MASt3R backbone predicts dense two-view geometry in the keyframe
coordinate system. Following the notation in~\cite{murai2025mast3r},
we denote this forward pass as
\begin{equation}
\mathrm{FM}(I_t, I_k) \rightarrow \{X_t^k,\; C_t^k\},
\end{equation}
where $X_t^k \in \mathbb{R}^{H \times W \times 3}$ is the dense
pointmap associated with frame $I_t$, and $C_t^k \in \mathbb{R}^{H \times W \times 1}$
is the corresponding confidence map. Rather than relying on sparse
feature landmarks or externally provided depth, MASt3R-SLAM uses
these dense geometric predictions as the intermediate representation
for tracking and local map construction.

To improve geometric consistency over time, MASt3R-SLAM incrementally
fuses the predicted pointmaps into a canonical local map attached to
the keyframe. Denoting the fused pointmap by
$\tilde{X}_k^k$ with confidence $\tilde{C}_k^k$, the local update can
be written as
\begin{equation}
\tilde{X}_k^k \leftarrow
\frac{\tilde{C}_k^k \tilde{X}_k^k + C_t^k X_t^k}
{\tilde{C}_k^k + C_t^k},
\qquad
\tilde{C}_k^k \leftarrow \tilde{C}_k^k + C_t^k.
\end{equation}
This fused dense geometry provides the geometric substrate used in our
pipeline. In the following sections, we augment this representation
with semantic information and further convert it into a local ESDF for
CBF-based safety filtering.

\subsection{Control Barrier Functions}

Control Barrier Functions (CBFs) provide a systematic tool for enforcing
safety constraints in control systems by ensuring forward invariance of a
safe set \cite{AmesTAC2017,Ames2019}.

Consider a control-affine continuous-time system
\begin{equation}
\dot{x} = f(x) + g(x)u,
\end{equation}
where $x \in \mathcal{X} \subset \mathbb{R}^n$ is the system state and $u \in \mathcal{U} \subset \mathbb{R}^m$
is the control input. Let $h : \mathbb{R}^n \rightarrow \mathbb{R}$ be a
continuously differentiable function that defines the safe set
\begin{equation}
\mathcal{C} = \{ x \mid h(x) \ge 0 \}.
\end{equation}
The function $h(x)$ is called a Control Barrier Function if there exists
an extended class-$\mathcal{K}$ function $\alpha(\cdot)$ such that for all
$x \in \mathcal{C}$,
\begin{equation}
\sup_{u \in \mathcal{U}} \left[ \mathcal{L}_f h(x) + \mathcal{L}_g h(x) u \right]
\ge -\alpha\big(h(x)\big),
\end{equation}
where $\mathcal{L}_f h(x) = \nabla h(x)^\top f(x)$ and
$\mathcal{L}_g h(x) = \nabla h(x)^\top g(x)$ denote the Lie derivatives.

In practice, a common choice is a linear class-$\mathcal{K}$ function
$\alpha(h) = \alpha h$ with $\alpha > 0$, leading to the condition
\begin{equation}
\mathcal{L}_f h(x) + \mathcal{L}_g h(x) u \ge -\alpha h(x),
\end{equation}
which ensures forward invariance of the safe set $\mathcal{C}$,
i.e., if $x(0) \in \mathcal{C}$, then $x(t) \in \mathcal{C}$ for all $t \ge 0$.

% \begin{remark}
% Although the CBF condition is formulated in continuous time, robotic
% systems operate in sampled-data settings where perception and control are
% updated at discrete time instants. In this work, the continuous-time CBF
% condition is enforced at each control step through a quadratic program (QP) (see Sec. \ref{subsec: CBF-QP}),
% using the latest state estimate and environment representation.
% \end{remark}

\section{METHODS}

\subsection{System Overview}
The proposed framework connects monocular perception, semantic-aware field construction, and CBF-based safe control in a unified online pipeline, as illustrated in Fig.~\ref{fig:system_overview}. Incoming monocular RGB frames are processed by two perception branches. The MASt3R-SLAM branch estimates camera motion and reconstructs dense 3-D geometry, while the semantic segmentation branch predicts per-pixel semantic labels from the same RGB stream. These two outputs are associated through the image-to-map correspondences from the SLAM front end, allowing semantic observations to be fused over time with the reconstructed geometry and forming a temporally stabilized geometric-semantic map.

The fused representation is then converted into a semantic-aware distance-field representation for safety evaluation. Specifically, reconstructed geometry is first integrated into a local Truncated Signed Distance Field (TSDF), which fuses repeated geometric observations and suppresses frame-level reconstruction noise. The TSDF surface support is converted into an obstacle-aware occupancy map using semantic filtering, and class-dependent inflation margins are applied to this occupancy map before Euclidean Signed Distance Field (ESDF) construction. This TSDF-to-occupancy-to-ESDF pipeline separates geometric fusion, semantic obstacle selection, and control-time distance querying, so semantic risk is encoded directly into the geometry of the distance field rather than introduced only as a post-hoc controller parameter. The resulting ESDF provides both the distance value and local gradient required by the safety filter.

At each control step, a reference command is generated either from the human operator in teleoperation or from a Control Lyapunov Function (CLF)-based nominal controller in autonomous navigation. The CBF-based Quadratic Program (QP) safe controller queries the semantic-aware ESDF, constructs the corresponding CBF constraint, and minimally modifies the reference command to produce a safe control input. 

% \begin{figure*}[t] 
%     \centering
%     \includegraphics[width=\linewidth]{figures/cbfslam_overview_v9.pdf}
% \caption{\small \textbf{Overview of the proposed online semantic-aware safe control framework.}
% Monocular RGB frames are processed by semantic segmentation and MASt3R-SLAM-based dense geometry estimation. Semantic labels are temporally fused with reconstructed 3D geometry, which is integrated into a local TSDF and converted into obstacle-aware occupancy before ESDF construction. Obstacle filtering and class-dependent inflation encode risk directly into the distance field. The resulting distance and gradient are used by a CBF-QP safety filter to minimally adjust the reference control and generate safe navigation or teleoperation commands.}
%     \label{fig:system_overview}
% \vspace{-3mm}
% \end{figure*}

\subsection{MASt3R-SLAM with Semantics}

\subsubsection{Semantic Integration}
Given an incoming frame $I_t$ and the current keyframe $I_k$, the MASt3R front-end predicts dense geometry in the keyframe coordinate frame as
% \begin{equation}
% F_{\mathrm{geo}}(I_t, I_k) \rightarrow \{X_t^k, C_t^k\},
% \end{equation}
$F_{\mathrm{geo}}(I_t, I_k) \rightarrow \{X_t^k, C_t^k\}$,
where $X_t^k \in \mathbb{R}^{H \times W \times 3}$ is the dense pointmap
and $C_t^k \in \mathbb{R}^{H \times W \times 1}$ 
% \textcolor{red}{Shuo: this is different from what you have for $C_t^k$ under eq. (1)}
is the associated confidence map.
In parallel, we apply EfficientViT~\cite{cai2023efficientvit} as a lightweight semantic segmentation network to predict a per-pixel semantic label map 
% \begin{equation}
% F_{\mathrm{sem}}(I_t) \rightarrow \mathcal{Y}_t,
% \end{equation}
$F_{\mathrm{sem}}(I_t) \rightarrow \mathcal{Y}_t$,
where $\mathcal{Y}_t \in \{1,\dots,C\}^{H \times W}$ denotes semantic labels over
$C$ classes.

To associate semantics with the reconstructed geometry, we reuse the
pixel correspondences already estimated by the SLAM tracker and fuse
current-frame semantic observations into the active keyframe map. 
Let $\ell_{\xi}^t$ denote the semantic label stored at keyframe-grid location $\xi$, and let $m_t(\xi)$ be its matched pixel in the current frame.
The semantic map is updated through a confidence-weighted temporal fusion rule of the form
\begin{equation}
\ell_{\xi}^{t+1}
=
\Phi\!\left(
\ell_{\xi}^{t},
\mathcal{Y}_t(m_t(\xi)),
q_{\xi}^{t}
\right),
\end{equation}
where $q_{\xi}^{t}$ is the matching confidence. Internally, $\Phi(\cdot)$ keeps a vote weight $w_{\xi}^{t}$ and updates it as
\begin{equation}
w_{\xi}^{t+1}
=
\begin{cases}
w_{\xi}^{t}+q_{\xi}^{t}, & \text{if the matched label agrees with } \ell_{\xi}^{t},\\
w_{\xi}^{t}-q_{\xi}^{t}, & \text{otherwise}.
\end{cases}
\end{equation}
The label is replaced only when this weight becomes negative. In this way, semantic information is attached to the dense reconstructed map with minimal additional computation and is then used by the downstream ESDF construction and safety filter.

\subsection{Semantic Map and ESDF Construction}
The semantic MASt3R-SLAM module produces a dense geometric map of the
environment together with semantic labels associated with the
reconstructed scene elements. To support safety-aware control, this
representation is converted into an ESDF.

Compared to raw point cloud representations, the ESDF provides a
spatially organized geometric field that enables more stable evaluation of both
distance and gradient. In point-cloud-based formulations, the distance
function is typically defined with respect to the nearest point, which
can lead to frequent switching of the closest point and discontinuities
in the gradient. In contrast, the ESDF yields a spatially more
consistent representation that is better suited for optimization-based
safety filtering. More importantly, it provides a convenient interface
for incorporating semantic information directly into the geometric
representation used by the controller.

% Starting from the reconstructed geometry, we first integrate the
% confidence-weighted pointmap observations into a local TSDF
% $\mathcal{T}$. 
Starting from the reconstructed geometry, we first integrate the
pointmap observations into a local TSDF $\mathcal{T}$, weighting each
observation by its SLAM matching confidence.
Each voxel stores a truncated signed distance value
updated from the current MASt3R-SLAM geometry and the camera pose, while
the associated semantic label is updated using the temporally fused
semantic observations. The TSDF is used as an intermediate fusion layer:
it aggregates temporally registered pointmap observations into a locally
coherent surface representation before the control-oriented ESDF is
built.

We then extract the surface support of the TSDF into a geometric
occupancy set $\mathcal{O}$ in the voxelized map. Let $\mathcal{S}$
denote the semantic labeling associated with these occupied regions. A
semantic-aware occupancy set is constructed as
\begin{equation}
\mathcal{O}_{\mathrm{sem}} = \Psi(\mathcal{O}, \mathcal{S}),
\end{equation}
where $\Psi(\cdot)$ denotes a semantic occupancy shaping operator. This
operator retains obstacle-relevant occupied regions and applies
semantic-dependent inflation to the selected occupied subset. In this
way, the TSDF provides temporally fused geometry, while the occupancy
map provides the discrete obstacle support on which semantic filtering
and inflation are applied before distance-field evaluation.

Specifically, let $\mathcal{L}_{\mathrm{obs}}$ denote the set of
semantic labels regarded as safety-relevant obstacles. Occupied regions
associated with labels outside $\mathcal{L}_{\mathrm{obs}}$ can be
excluded from the semantic-aware occupancy representation. For occupied
regions associated with labels in $\mathcal{L}_{\mathrm{obs}}$, we apply
a semantic-dependent safety margin. Denoting this margin by
$d_{\mathrm{safe},l}$ for label group $l$, the resulting occupancy set
becomes more conservative around semantically important objects while
remaining less restrictive around low-risk categories.

The ESDF is then computed from the semantic-aware occupancy set:
\begin{equation}
d(x)=\mathrm{ESDF}(\mathcal{O}_{\mathrm{sem}})(x),
\end{equation}
where $d(x)$ represents the signed distance from query position $x$ to
the closest occupied region after semantic shaping. This distance field
is later used to construct the CBF constraint for safety filtering. We
construct a single semantic-aware ESDF rather than one ESDF per semantic
class, which keeps the control-time query to a standard grid lookup and
helps preserve online computational efficiency.

\begin{algorithm}[!h]
\caption{Semantic-Aware ESDF Construction}
\label{alg:semantic_esdf}
\begin{algorithmic}[1]
\Require Reconstructed pointmaps with poses and confidences, semantic labels $\mathcal{S}$, obstacle label set $\mathcal{L}_{\mathrm{obs}}$, safety margins $\{d_{\mathrm{safe},l}\}$
\Ensure Distance field $d(\cdot)$

\State Integrate reconstructed geometry into a local TSDF $\mathcal{T}$
\State Extract surface-supported occupancy $\mathcal{O}$ from $\mathcal{T}$
\State Initialize semantic-aware occupancy $\mathcal{O}_{\mathrm{sem}} \gets \emptyset$
\For{each occupied element $o \in \mathcal{O}$}
    \State obtain its semantic label $l$ from $\mathcal{S}$
    \If{$l \in \mathcal{L}_{\mathrm{obs}}$}
        \State inflate $o$ according to $d_{\mathrm{safe},l}$
        \State insert the inflated region into $\mathcal{O}_{\mathrm{sem}}$
    \EndIf
\EndFor
\State compute ESDF from $\mathcal{O}_{\mathrm{sem}}$
\State $d(x) \gets \mathrm{ESDF}(\mathcal{O}_{\mathrm{sem}})(x)$
\Return $d(\cdot)$
\end{algorithmic}
\end{algorithm}
\vspace{-3mm}

\subsection{CBF Formulation with Semantics}

Using the distance field obtained from the ESDF, we
construct a CBF that encodes safety
constraints with respect to nearby obstacles.

Let $d(x)$ denote the ESDF value at position $x$, which
incorporates class-dependent safety margins through
semantic-aware map construction. 
% The safety function is
% defined as
% \begin{equation}
% h(x) = d(x).
% \end{equation}

% The safe set induced by the constructed CBF is given by
% \begin{equation}
% \mathcal{C} = \{ x \mid d(x) \ge 0 \}.
% \end{equation}
We define the safety function as \(h(x)=d(x)\), which induces the safe
set \(\mathcal{C}=\{x\mid d(x)\ge 0\}\).
To ensure forward invariance of $\mathcal{C}$ under
continuous-time dynamics, we impose the standard CBF
condition
\begin{equation}\label{eq:cbf_semantic}
\mathcal{L}_f h(x) + \mathcal{L}_g h(x) u \ge -\alpha_l h(x),
\end{equation}
where $\alpha_l > 0$ is a class-dependent CBF gain selected according to the
semantic class of the closest obstacle.

In this formulation, semantic information influences the
safety filter in two complementary ways. First, semantic
labels shape the geometry of the ESDF through
class-dependent inflation, which modifies the underlying
safety function $h(x)$. Second, semantics modulate the
barrier response through $\alpha_l$, allowing different
levels of conservativeness depending on obstacle type.

\begin{remark}
We assume that the semantic-dependent CBF parameters,
including the class-$\mathcal{K}$ gains $\alpha_l$ and
safety margins $d_{\mathrm{safe},l}$, are available
\emph{a priori}. These parameters can be obtained through
data-driven or learning-based approaches such as
BarrierNet~\cite{xiao2023barriernet},
reinforcement-learning-based CBF tuning~\cite{ehsan2024rlcbf},
graph neural network methods~\cite{gao2023gnn-cbf}, or
vision-language models~\cite{ChenChandra2026}.
Our contribution focuses on how semantic information is
incorporated into the geometric representation and propagated to the resulting CBF constraints
at runtime.
\end{remark}

\subsection{Optimization-Based Safety Filter}
\label{subsec: CBF-QP}

The safety condition defined above is enforced through a
QP-based safety filter. At each control
step, the controller computes a control input that remains
as close as possible to a given reference command while
satisfying the CBF constraint.
For clarity, we present the formulation for first-order
control-affine dynamics, which directly applies to
velocity-controlled systems used in our hardware experiments.
The formulation naturally extends to higher-order systems,
as discussed below.

Given the semantic CBF condition in~\eqref{eq:cbf_semantic}, the safety
constraint can be written as an affine constraint in the control input $u$:
\begin{equation}
\nabla h(x)^\top \big(f(x) + g(x)u\big) \ge -\alpha_l h(x),
\end{equation}
where $\nabla h(x)=\nabla d(x)$ is evaluated from the ESDF. For
higher-order systems, the same construction is applied to higher-order
barrier derivatives, leading to affine constraints in which the control
appears through the corresponding Lie-derivative term, while the
remaining terms are state-dependent. In our implementation, ESDF
gradients and Hessians are computed by finite differences and evaluated
at query points using trilinear interpolation, yielding constraints that
can be directly incorporated into the QP-based safety filter.

% At each control step, the safe control input is obtained by
% solving the following optimization problem:
% \begin{equation}
% \begin{aligned}
% \min_{u} \quad & \tfrac{1}{2}\|u - u_{\mathrm{ref}}\|^2 \\
% \text{s.t.} \quad & A(x) u \ge b(x), ~u\in \mathcal{U},
% \end{aligned}
% \end{equation}
% \textcolor{red}{Shuo: the cost looks different from the one in Fig. 2: position of 2, input bounds}where $A(x) = \nabla h(x)^\top g(x)$, $b(x) = -\alpha_l h(x) - \nabla h(x)^\top f(x)$.

At each control step, the safe control input is obtained by
solving the following optimization problem, which expands the
schematic CBF-QP block in Fig.~\ref{fig:system_overview} with the
conventional objective scaling and explicit input bounds:
\begin{equation}
\begin{aligned}
\min_{u\in \mathcal{U}} \quad & \tfrac{1}{2}\|u - u_{\mathrm{ref}}\|_2^2 \\
\text{s.t.} \quad & \mathcal{L}_f h(x) +\mathcal{L}_g h(x)u \ge -\alpha_l h(x).
\end{aligned}
\end{equation}
The proposed safety filter operates on a reference control 
input $u_{\mathrm{ref}}$ that is generated by different sources depending on
the task. In teleoperation, the reference input is directly
provided by the human operator. In autonomous navigation,
a CLF-based controller is used
to generate a goal-directed nominal input following~\cite{coffey2023reactive}. The CBF-QP then modifies this reference only when necessary to ensure safety.

\begin{remark}
The ESDF is a sampled distance field and is not globally smooth due to
obstacle switching and map discretization. We therefore construct the
CBF constraint using the local ESDF value and its numerically evaluated
gradient at the current query point. This is consistent with
distance-field-based CBF methods, where local distance and gradient
queries yield effective safety constraints despite the non-smoothness of
the underlying representation~\cite{zhou2025control,Raja2024OGMCBF}.
\end{remark}

\section{Validation}
\label{sec: case study}
This section evaluates the proposed semantic-aware safety framework in both simulation and physical hardware experiments. We demonstrate the computational efficiency and risk-allocation advantages of our approach against baselines using a simulated benchmark. Subsequent teleoperation and autonomous navigation experiments assess the system's capacity for online, semantic-driven collision avoidance on real-world hardware.
\vspace{-3mm}
\subsection{Simulation}

We conduct simulations to evaluate the computational efficiency and the semantic-aware safety behavior of our proposed method. We compare three methods:
(i) SaferSplat \cite{chen2024safer-splat}, a recent Gaussian-splatting-based method for safety filtering;
(ii) Ours (ESDF), which uses the same ESDF-based safety filter without
semantic consideration; and (iii) Ours (Semantic ESDF), which incorporates
semantic-dependent obstacle inflation into the ESDF construction.

\begin{table}[h]
\centering
\caption{\small Equal-time comparison on the benchmark. Matched-progress clearance and collision rate are evaluated only up to the common progress reached by all methods on the same trajectory.}
\label{tab:sim_3methods}
\setlength{\tabcolsep}{3pt}
\resizebox{\columnwidth}{!}{
\begin{tabular}{lcccc}
\toprule
Method 
& \makecell{Computation\\Time (s) $\downarrow$} 
& \makecell{Progress\\to Goal $\uparrow$} 
& \makecell{Matched Progress\\Clearance (m) $\uparrow$} 
& \makecell{Matched Progress\\Collision Rate $\downarrow$} \\
\midrule
SaferSplat & 0.149 & 0.360 & 0.101 & 0.216 \\
Ours (ESDF) & 0.002 & 0.697 & 0.106 & 0.218 \\
Ours (Semantic ESDF) & 0.002 & 0.603 & 0.108 & 0.200 \\
\bottomrule
\end{tabular}
} 
\end{table}

\begin{table*}[t]
\centering
\caption{\small Risk-group ablation on the benchmark with 486 matched trajectories. Matched progress metrics evaluate safety only up to the common progress reached by both ESDF variants on the same trajectory.}
\label{tab:risk_ablation}
\begin{tabular}{lcccccc}
\toprule
& \multicolumn{3}{c}{Ours (ESDF)} & \multicolumn{3}{c}{Ours (Semantic ESDF)} \\
\cmidrule(lr){2-4} \cmidrule(lr){5-7}
Risk Group 
& \makecell{Progress\\to Goal $\uparrow$}
& \makecell{Matched Progress\\Clearance (m) $\uparrow$} 
& \makecell{Matched Progress\\Collision Rate $\downarrow$} 
& \makecell{Progress\\to Goal $\uparrow$} 
& \makecell{Matched Progress\\Clearance (m) $\uparrow$} 
& \makecell{Matched Progress\\Collision Rate $\downarrow$} \\
\midrule
High & 0.669 & 0.128 & 0.146 & 0.534 & 0.134 & 0.116 \\
Mid  & 0.777 & 0.092 & 0.248 & 0.631 & 0.093 & 0.224 \\
Low  & 0.647 & 0.098 & 0.261 & 0.647 & 0.098 & 0.261 \\
\bottomrule
\end{tabular}\vspace{-3mm}
\end{table*}
\subsubsection{Experimental Setup}
We evaluate the proposed framework on a benchmark containing six scenes of ScanNet++ dataset ~\cite{yeshwanthliu2023scannetpp}
using 486 matched start-goal trajectories generated from an object-centric
sampling protocol.
All methods are evaluated on identical trajectories within each scene
to ensure a fair comparison.
To analyze semantic effects, object classes are grouped into high-, mid-,
and low-risk categories based on their scene-level geometric scale,
while removing structural elements (e.g., walls and floors). This grouping
is used in the semantic ablation study to assess how safety behavior varies
across different risk levels.

The robot is modeled as a double-integrator system with
\(\dot p=v,\ \dot v=u\), where \(p,v\in\mathbb{R}^3\)
denote position and velocity, respectively, and
\(u\in\mathbb{R}^3\) is the control input.
% The robot is modeled as a double-integrator system
% \begin{equation}
% \dot{p}=v,\qquad \dot{v}=u,
% \end{equation}
% where $p\in\mathbb{R}^3$ and $v\in\mathbb{R}^3$ denote position and velocity,
% and $u\in\mathbb{R}^3$ is the control input. 
The corresponding high-order CBF constraint takes the form
\begin{equation}
\nabla d(p)^\top u
+ v^\top \nabla^2 d(p)\,v
+ \alpha_{l,1}\nabla d(p)^\top v
+ \alpha_{l,2}\, d(p)
\ge 0,
\end{equation}
where $\alpha_{l,1},\alpha_{l,2}>0$ are high-order CBF gains. 

All methods use the same nominal goal-reaching controller and identical
CBF gains. To ensure a consistent comparison under online constraints,
each method is given the same time budget to incrementally construct
its scene representation. SaferSplat builds a continuously optimized
3-D Gaussian Splatting map, while our ESDF baseline constructs a distance
field from the same input stream. The semantic ESDF variant further
applies class-dependent obstacle inflation, with margins of 0.15\,m,
0.05\,m, and 0.00\,m for high-, mid-, and low-risk objects, respectively.

\subsubsection{Metrics}
We evaluate performance using four metrics: computation time,
progress to goal, matched progress (MP) clearance to mesh, and
MP collision rate.
Computation time is the average per-step runtime of the safety filter.
Progress to goal measures the normalized distance traveled toward the
goal. Since different safety filters may reach different portions of the
same trajectory, we compute clearance-related metrics over the common
progress reached by the compared methods. MP clearance is defined as
the smallest signed distance to the ground-truth mesh after subtracting
the robot radius within this common-progress segment. MP collision rate
is the fraction of trajectories with negative MP clearance. This protocol
prevents a conservative method from receiving an artificially favorable
clearance score simply because it stops before reaching more difficult
parts of the path. Therefore, MP clearance and MP collision rate are
interpreted jointly with progress to goal in Tables~\ref{tab:sim_3methods}
and~\ref{tab:risk_ablation}.

\begin{figure}[!ht] 
    \centering
    \vspace{-5mm}
    \subfloat[Teleoperation with a ball.]{
        \includegraphics[width=0.45\columnwidth,trim={0 0 0 20mm},
            clip]{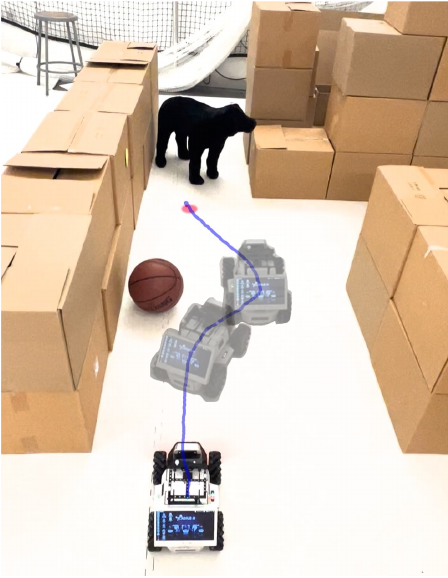}
        
        \label{fig:tele_ball}
    }
    \hspace{-3mm}
    \subfloat[Teleoperation with a dog.]{
        \includegraphics[width=0.45\columnwidth,trim={0 0 0 20mm},
            clip]{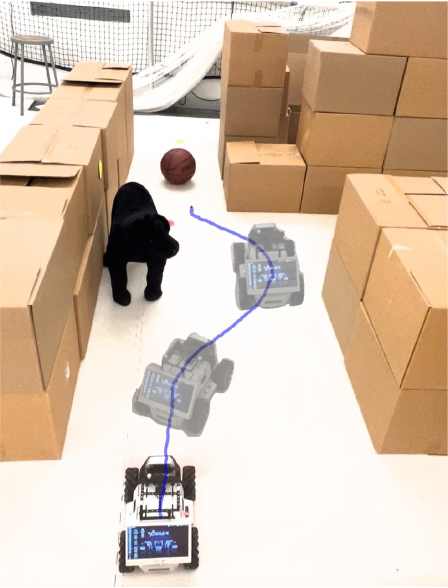}
        \label{fig:tele_dog}
    }
    \caption{\small Teleoperated robot trajectories under different obstacle semantics. (a) Low-risk obstacle (ball), where the robot allows closer interaction with minimal intervention. (b) High-risk obstacle (dog), where the safety filter activates earlier and maintains a larger clearance.}
    \label{fig:teleoperation}
\vspace{-5mm}
\end{figure}

\begin{figure*}[!t]
\centering
\setlength{\tabcolsep}{0pt}
\begin{tabular}{@{}m{0.015\linewidth}@{\hspace{2mm}}m{0.95\linewidth}@{}}
\centering\raisebox{10mm}{\rotatebox[origin=c]{90}{\small \textbf{Ball}}} &
\includegraphics[width= 0.95\linewidth]{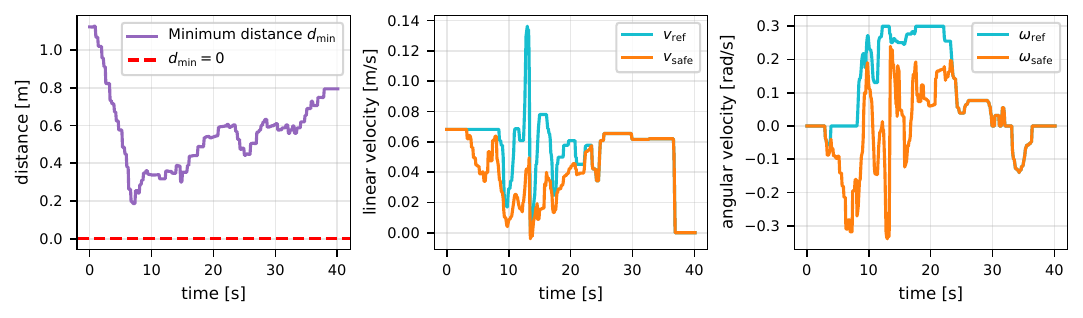}
\end{tabular}

\vspace{-3mm}

\begin{tabular}{@{}m{0.015\linewidth}@{\hspace{2mm}}m{0.95\linewidth}@{}}
\centering\raisebox{10mm}{\rotatebox[origin=c]{90}{\small \textbf{Dog}}} &
\includegraphics[width=0.95\linewidth]{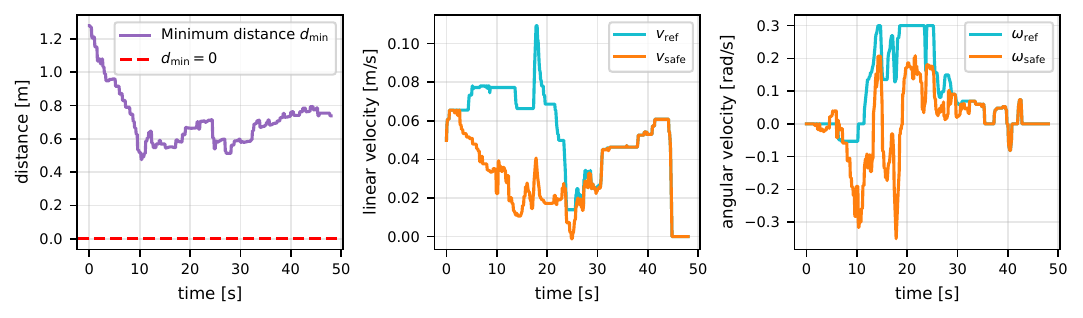}
\end{tabular}
\vspace{-3mm}

\caption{\small Teleoperation results with semantic CBF-based safety filtering. Each row shows the obstacle distance (left), linear
velocity (middle), and angular velocity (right). The first row corresponds to the low-risk ball, where the robot approaches with smaller clearance and the safety filter intervenes later. The second row corresponds to the high-risk dog, where the safety filter activates earlier and
maintains a larger distance.}
\label{fig:semantic_tele}
\vspace{-5mm}
\end{figure*}

\subsubsection{Comparison Results}
Table~\ref{tab:sim_3methods} reports the equal-time comparison on the benchmark. SaferSplat achieves substantially lower progress
to goal under the same time budget, indicating more conservative and
computationally heavier behavior. After matching progress across
methods, its clearance advantage from stopping earlier is removed, and
the ESDF-based methods achieve comparable or better MP clearance while
maintaining substantially higher task progress. Both ESDF-based methods
also operate roughly 75 times faster. This efficiency comes
from the representation used at control time: after the local map is
converted to an ESDF, each safety-filter update requires only grid-based
distance and gradient queries followed by a small CBF-QP. SaferSplat,
by contrast, relies on a continuously optimized Gaussian-splatting scene
representation and an ellipsoid-based safety model, which requires more
expensive geometric evaluation when constructing the safety constraint.
The lower runtime of the ESDF query also helps the controller preserve
goal-directed motion under the same online budget, explaining the
higher progress-to-goal values. The semantic ESDF variant has slightly
lower progress than the plain ESDF because high- and mid-risk obstacles
are intentionally inflated, causing earlier intervention near these
objects rather than a loss of computational efficiency.
Compared with the plain ESDF controller, the semantic-aware ESDF
variant preserves the same runtime and improves MP clearance,
demonstrating that semantic shaping enhances safety without sacrificing
computational efficiency.

Table~\ref{tab:risk_ablation} further isolates the effect of
semantic information within our framework. The semantic-aware controller
selectively increases conservativeness for high- and mid-risk objects,
while leaving low-risk behavior largely unchanged. For high-risk targets,
the MP clearance increases from 0.128\,m to 0.134\,m and the MP
collision rate decreases from 0.146 to 0.116, while low-risk
trajectories remain identical. These results show that the proposed
method does not uniformly increase conservativeness, but
reallocates safety margins according to semantic risk, which is the
intended behavior of the framework.

\subsection{Hardware Experiment}

We validate the proposed framework in both teleoperation and
autonomous navigation scenarios to demonstrate its effectiveness
for semantic-aware safe control in an online setting. Hardware
experiments are conducted on a desktop computer (Ubuntu 20.04)
equipped with an RTX~3080 GPU and implemented within the Robot
Operating System (ROS). The resulting QP
is solved using \texttt{CVXPY} \cite{diamond2016cvxpy}. Under this setup, the full pipeline runs at 10\,Hz and is primarily limited by GPU computation, while separate evaluation of the perception-only module on an RTX~4090 achieves up to 20\,Hz. Recorded demonstrations of the teleoperation and
navigation hardware experiments are provided in the supplementary video.

In the teleoperation task, the human operator uses a Logitech Extreme
3D Pro joystick
% \footnote{https://www.logitechg.com/en-us/shop/p/extreme-3d-pro-joystick}
to steer an AgileX LIMO
% \footnote{https://global.agilex.ai/products/limo}  
robot through the environment.
We impose the CBF constraint at a look-ahead point
located a distance \(L\) along the robot heading. Let
\(x=[p_x,p_y,\theta]^\top\) denote the robot pose and
\(u=[v,\omega]^\top\) denote the commanded linear and angular
velocities. The look-ahead point dynamics can be expressed as
\begin{equation}
\dot {p}_L =
\begin{bmatrix}
\cos\theta & -L\sin\theta\\
\sin\theta &  L\cos\theta
\end{bmatrix}
\begin{bmatrix}
v\\
\omega
\end{bmatrix}.
\end{equation}
At each control step, the safe input is obtained by solving
\begin{equation}
\begin{aligned}
\min_{v,\omega}\quad &
\lambda_v(v-v_{\mathrm{ref}})^2
+\lambda_{\omega}(\omega-\omega_{\mathrm{ref}})^2 \\
\text{s.t.}\quad &
\nabla d(p_L)^\top
\begin{bmatrix}
\cos\theta & -L\sin\theta\\
\sin\theta &  L\cos\theta
\end{bmatrix}
\begin{bmatrix}
v\\
\omega
\end{bmatrix}
\ge -{\alpha_l}d(p_L),
\end{aligned}
\end{equation}
where \(d(p_L)\) and \(\nabla d(p_L)\) are obtained from the
semantic-aware ESDF, \(\alpha_l\) is the class-dependent CBF gain, and
\(\lambda_v,\lambda_\omega>0\) penalize deviations from the reference
command.

% In the teleoperation task, the human operator uses a Logitech Extreme 3D Pro Joystick to steer an AgileX LIMO
% % \footnote{https://www.logitechg.com/en-us/shop/p/extreme-3d-pro-joystick}
% % \footnote{https://global.agilex.ai/products/limo}
% robot through a remote environment. We model the robot as a unicycle system with linear
% and angular velocity inputs:
% \begin{equation}
% \dot{p}_x = v\cos\theta,\quad
% \dot{p}_y = v\sin\theta,\quad
% \dot{\theta} = \omega,
% \end{equation}
% where $x=[p_x,p_y,\theta]^\top$ denotes the robot state and
% $u=[v,\omega]^\top$ is the control input.
% Safety is evaluated at a look-ahead point located at a distance
% $L$ along the heading direction from the geometric center of the robot.
% Its dynamics can be written as
% \begin{equation}
% \dot p_L =
% \begin{bmatrix}
% \cos\theta & -L\sin\theta\\
% \sin\theta &  L\cos\theta
% \end{bmatrix}
% \begin{bmatrix}
% v\\
% \omega
% \end{bmatrix}.
% \end{equation}
% At each control step, the safe input is obtained by solving the
% following QP:
% \begin{equation}
% \begin{aligned}
% \min_{v,\omega}\quad &
% \lambda_{v}(v-v_{\mathrm{ref}})^2
% +\lambda_{\omega}(\omega-\omega_{\mathrm{ref}})^2 \\
% \text{s.t.}\quad &
% \nabla d(p_L)^\top
% \begin{bmatrix}
% \cos\theta & -L\sin\theta\\
% \sin\theta &  L\cos\theta
% \end{bmatrix}
% \begin{bmatrix}
% v\\
% \omega
% \end{bmatrix}
% \ge -{\alpha_l}d(p_L),
% \end{aligned}
% \end{equation}
% where $\lambda_{v}$ and $\lambda_{\omega}$ are scalar coefficients that penalize the deviation of the linear and angular velocities from the reference command.

\subsubsection{\textbf{Teleoperation}} 

We first evaluate how semantic information influences the
safety behavior during human-in-the-loop control. In this
experiment, a LIMO robot is teleoperated to avoid two types
of obstacles: a basketball (low-risk) and a dog (high-risk),
as shown in Fig.~\ref{fig:tele_ball} and~\ref{fig:tele_dog}.
Each task is repeated 12 times to account for variability in human inputs, and average values are reported. We measure the first intervention distance and the minimum
clearance distance.
The quantitative results show a clear semantic-dependent
behavior. For the low-risk obstacle (basketball), the average
first intervention distance is $0.36$ m and the minimum
clearance is $0.14$ m. In contrast, for the high-risk obstacle
(dog), the intervention distance increases to $0.56$ m and
the minimum clearance to $0.41$ m.
This difference is also reflected in the example plots in
Fig.~\ref{fig:semantic_tele}. As shown in the results from the first row, the robot approaches the
basketball with a smaller safety margin, resulting in a
lower minimum distance. In contrast, the robot maintains a
consistently larger distance when avoiding the dog, and the safety filter activates earlier. The corresponding control inputs further support this
observation. The velocity profiles in Fig.~\ref{fig:semantic_tele}
show that the robot undergoes earlier and stronger
modifications when approaching the high-risk obstacle,
indicating a more conservative response induced by semantic
information. 

\subsubsection{\textbf{Navigation}}

\begin{figure}[!t] 
    \centering
    \subfloat[Navigation in a loop.]{
        \includegraphics[width=0.46\columnwidth]{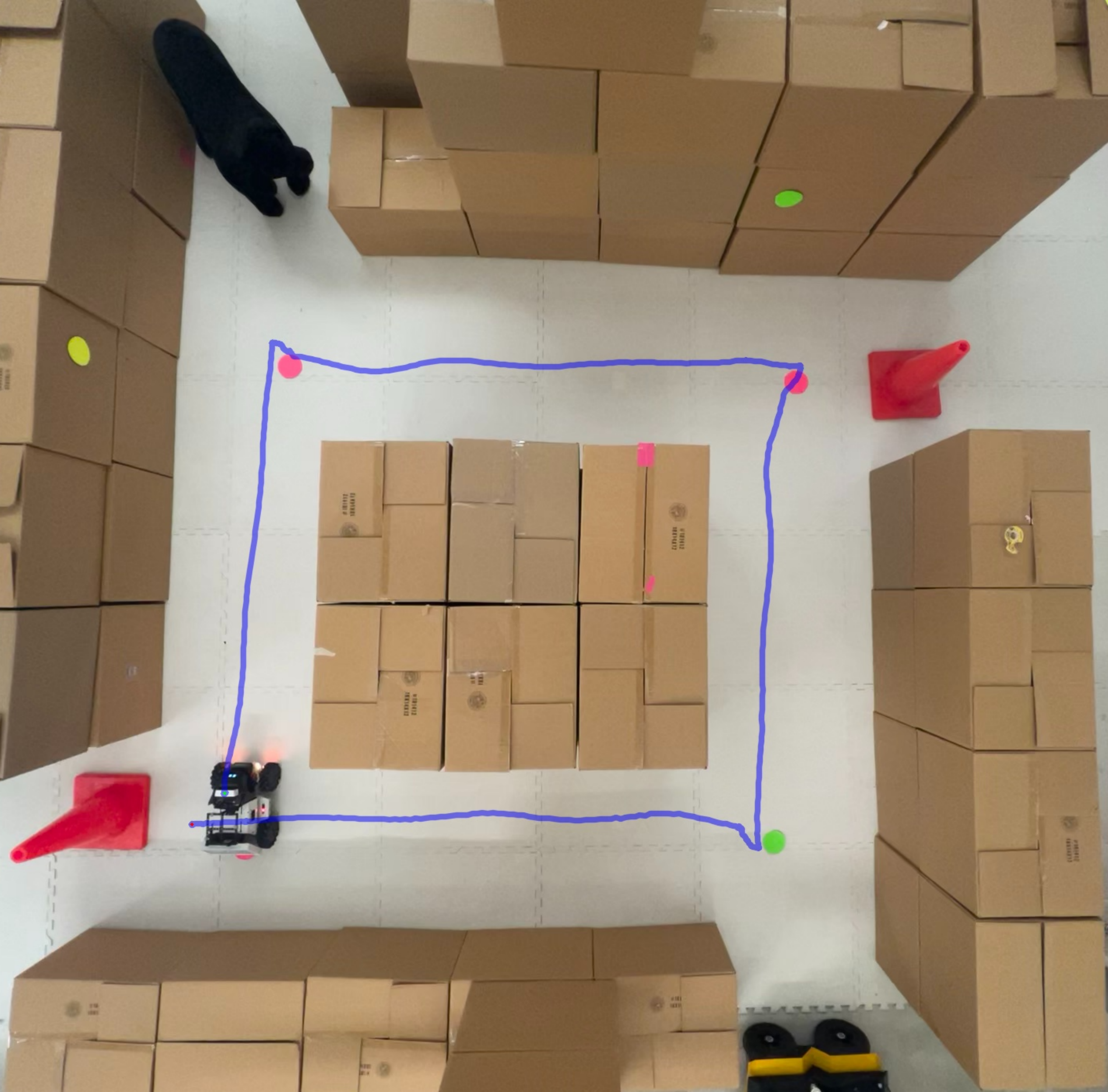}
        \label{fig:loop}
    }
    \hspace{-2mm}
    \subfloat[Navigation with unknown obstacle.]{
        \includegraphics[width=0.46\columnwidth]{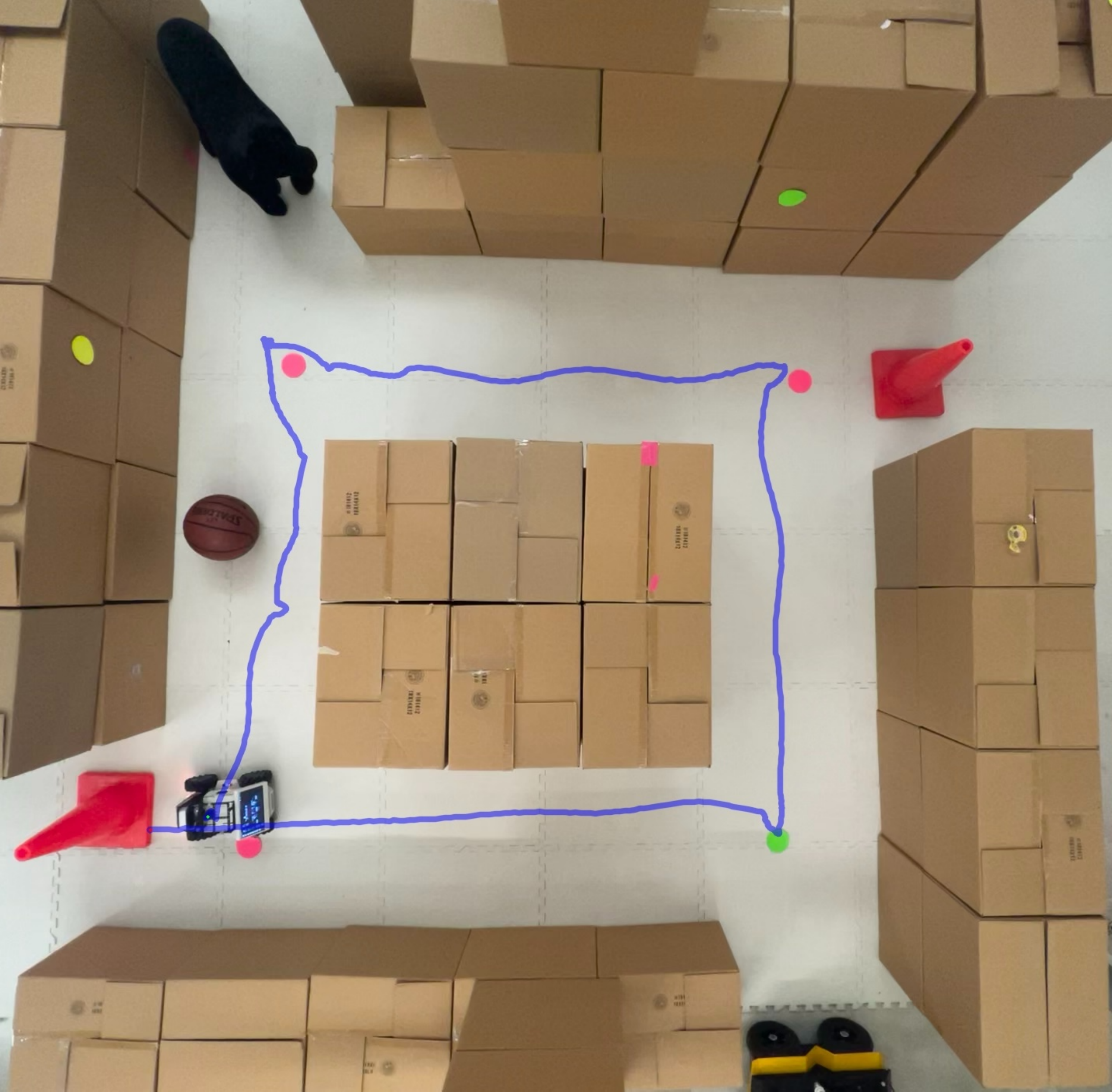}
        \label{fig:loop_with_ball}
    }
    \caption{\small Overhead view of the navigation experiment. The robot starts from the lower-left corner. (a) The robot completes a loop using the RGB-only SLAM map. (b) An unseen obstacle is introduced along the path, and the robot updates the local ESDF, deviates to avoid it, and continues the task.}
    \label{fig:navigation}
\vspace{-5mm}
\end{figure}
We further evaluate the proposed framework in an autonomous
navigation task to demonstrate its integration with the RGB-only
SLAM pipeline and its ability to perform online collision
avoidance, as summarized in Fig.~\ref{fig:navigation}. The experiment consists of two phases. In the first phase, a
human operator teleoperates the robot along a rectangular loop
to build an initial map. The robot then autonomously completes
the loop by sequentially reaching  the goal waypoint at each corner of the loop, as shown in
Fig.~\ref{fig:loop}.
In the second phase, an unseen obstacle (a basketball) is
placed on the robot’s path. The robot is required to complete
the same loop again while adapting to the newly introduced
obstacle. As shown in Fig.~\ref{fig:loop_with_ball}, the robot
successfully deviates from the nominal trajectory to avoid
the unseen obstacle and then returns to the original path to
complete the loop. This demonstrates that the proposed
framework enables online collision avoidance while
maintaining consistent navigation behavior.
Notably, the avoidance maneuver is triggered without prior
knowledge of the obstacle, highlighting the ability of the
framework to incorporate newly perceived obstacles into the CBF constraints in an online manner.

\section{Conclusion and future work}
This letter presents a semantic-aware CBF-based
framework for safe robotic navigation and teleoperation,
which integrates RGB-only perception (via foundation-model-based SLAM and semantic segmentation), ESDF construction,
and optimization-based safety filtering within a unified
pipeline. Semantic information is incorporated into both the distance-field geometry and the CBF constraint, enabling semantic-aware safety behavior. The proposed method is validated in both simulation and hardware experiments. Results show that the framework
achieves efficient online performance while adapting safety
behavior according to obstacle semantics, and enables
online collision avoidance with previously unseen
obstacles. 
Future work will focus on incorporating uncertainty-aware
perception into the CBF framework, extending the approach
to dynamic environments, and improving the robustness and
scalability of the system for real-world deployment.

\bibliographystyle{IEEEtran}
\bibliography{biblio/1_cbf,biblio/2_vision, biblio/3_others}

\end{document}